%% file: acl_latex.tex
\pdfoutput=1

\documentclass[11pt]{article}

\usepackage[final]{acl}

\usepackage{times}
\usepackage{latexsym}

\usepackage[T1]{fontenc}

\usepackage[utf8]{inputenc}

\usepackage{microtype}

\usepackage{inconsolata}

\usepackage{graphicx}
\usepackage{subfig}
\usepackage{graphicx}
\usepackage{float}
\usepackage{array}
\usepackage{colortbl}
\usepackage{inconsolata}
\usepackage{microtype}
\usepackage{times}
\usepackage{pifont}
\usepackage{url}
\usepackage{latexsym}
\usepackage{microtype}
\usepackage{booktabs}
\usepackage{amsmath}
\usepackage{amsthm}
\usepackage{enumitem}
\usepackage{color,amsfonts}
\usepackage{multirow,array}
\usepackage{tikz}
\usepackage{pgfplots}
\pgfplotsset{compat=1.18}
\usepackage{algpseudocode}
\usepackage{amssymb,mathtools}
\usepackage[linesnumbered,ruled,vlined]{algorithm2e}
\usepackage{amssymb}
\usepackage{nomencl}
\usepackage{etoolbox}
\usepackage{tcolorbox}
\theoremstyle{definition}
\usepackage{hyperref}
\usepackage{cleveref}
\usepackage{multirow}
\usepackage{xspace}

\newcommand{\datasetNameObj}[0]{Spatial-Obj\xspace}
\newcommand{\datasetNameCoT}[0]{Spatial-CoT\xspace}
\newcommand{\LMMs}[0]{LMMs\xspace}

\title{An Empirical Analysis on Spatial Reasoning Capabilities of \\ Large Multimodal Models}

\author{Fatemeh Shiri$^{\heartsuit}$ \enskip Xiao-Yu Guo$^{*\diamondsuit}$ \enskip Mona Golestan Far$^\heartsuit$ \\ \textbf{Xin Yu$^\clubsuit$ \enskip Gholamreza Haffari$^\heartsuit$ \enskip Yuan-Fang Li$^\heartsuit$} \\
  $^\heartsuit$Department of Data Science \& AI, Monash University \\
  $^\diamondsuit$ Australian Institute for Machine Learning, University of Adelaide\\
  $^\clubsuit$School of Electrical Engineering and Computer Science, University of Queensland\\
    \texttt{\{first.last\}@monash.edu}\\
    \texttt{xiaoyu.guo@adelaide.edu.au }\\
    \texttt{xin.yu@uq.edu.au}
}

\begin{document}
\maketitle
\begin{abstract}
\input{sections/00-abstract}
\end{abstract}

\input{sections/01-Introduction}

\input{sections/02-Related-Work}

\input{sections/03-Spatial-MM-Benchmark}
\input{sections/04-Methodology}
\input{sections/05-Experiment}
\input{sections/06-Conclusion}
\input{sections/limitations}
\section*{Acknowledgments}
This material is partially supported by the DARPA Assured Neuro Symbolic Learning and Reasoning (ANSR) program under award number FA8750-23-2-1016.

\bibliography{spatial-mm}

\clearpage

\appendix

\input{sections/appendix}

\end{document}

%% file: sections/00-abstract.tex
Large Multimodal  Models (LMMs) have achieved strong performance across a range of vision and language tasks. 
However, their \emph{spatial} reasoning capabilities are under-investigated. In this paper, we construct a novel VQA dataset, Spatial-MM, to comprehensively study LMMs' spatial understanding and reasoning capabilities. 
Our analyses on object-relationship and multi-hop reasoning reveal several important findings. 
Firstly, bounding boxes and scene graphs, even synthetic ones, can significantly enhance LMMs' spatial reasoning. 
Secondly, LMMs struggle more with questions posed from the human perspective than the camera perspective about the image. 
Thirdly, chain of thought (CoT) prompting does not improve model performance on complex multi-hop questions involving spatial relations. 
Lastly, our perturbation analysis on GQA-spatial reveals that LMMs are much stronger at basic object detection than complex spatial reasoning.  
We believe our benchmark dataset and in-depth analyses can spark further research on LMMs spatial reasoning. \footnote{Spatial-MM benchmark is available at: \url{https://github.com/FatemehShiri/Spatial-MM}}

%% file: sections/01-Introduction.tex
\section{Introduction}

Large Multimodal Models (\LMMs) have shown impressive generalization ability on several vision and language tasks. Several recent works, however, showed that these models lack spatial understanding \cite{eyes_wide,seed_bench,mm_bench, lei2024scaffolding,prasad2023rephrase}. As can be seen in Figure \ref{fig:open}, multimodal LLMs, including GPT-4o, often fail to answer questions from a human perspective within an image.
The focus of this work is to study the understanding of \emph{spatial} relations by top-performing \LMMs. Moreover, we go beyond evaluating only the final answers to directly analyzing the intermediate reasoning steps generated by chain of thought (CoT) prompting in multi-hop visual question-answering (VQA) tasks. We ground \LMMs’ reasoning steps into a scene graph format and verify whether they form a valid path. 

More specifically, we ask the following questions: 
(i)  What spatial relations are missed by models, and why it happen? 
(ii)  How can additional symbolic visual information, such as bounding boxes or scene graphs, improve the performance of \LMMs? Which of these symbolic information is more useful, and how can they be integrated in the reasoning process effectively? 
(iii)   How does the questions complexity affect \LMMs in handling spatial relations? 
(iv) How does the reasoning path of \LMMs  behave when they fail to answer a multi-hop question? Is the failure due to incorrect spatial reasoning or non-spatial reasoning?

\input{figures/open}

To address these questions, we construct Spatial-MM, a novel, challenging dataset, and comprehensively \LMMs spatial reasoning capabilities from different angles.
We analyze four top-performing \LMMs on Spatial-MM and GQA-spatial \cite{whatsup} benchmarks to identify problems with visual question answering (VQA) evaluation methodology.
Our comprehensive analyses reveal a number of important insights that point to future research directions. Our contributions can be summarized as follows. 

\noindent$\bullet$ We present a new, challenging spatial-aware benchmark that incorporates a variety of spatial relationship types, accounting for both human and camera perspectives. 

\noindent$\bullet$ Our coprehensive empirical analyses show that: (i) bounding boxes and scene graphs, even synthetic ones, can significantly enhance LMMs' spatial reasoning, (ii) LMMs struggle more with questions posed from the human perspective than the camera perspective about the image, (iii) chain of thought (CoT) prompting does not improve model performance on complex multi-hop questions involving spatial relations, and (iv)  LMMs are much stronger at basic object detection than complex spatial reasoning.  

%% file: figures/open.tex
\begin{figure}[t]
    \centering
    \resizebox{.5\textwidth}{!}{
    \includegraphics{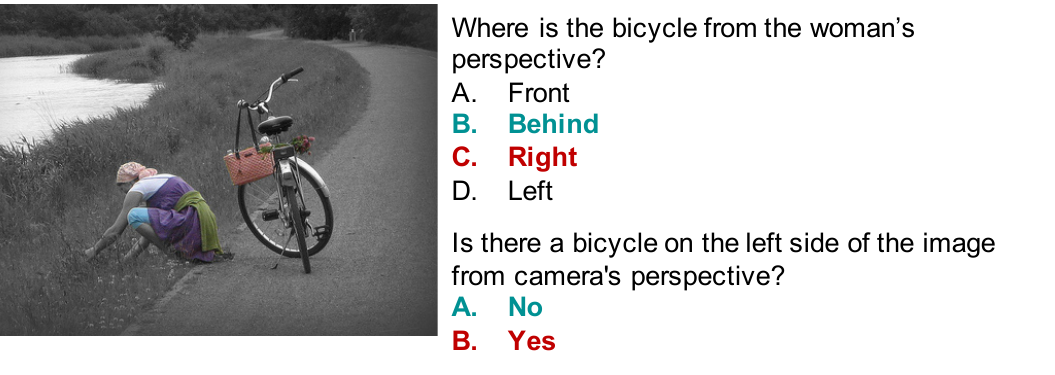}}
    \caption{benchmarking the spatial reasoning capabilities of GPT-4o \cite{gpt4} (Date accessed: June 12, 2024). Text in \textcolor{red}{red} and \textcolor{green}{green} signifies an incorrect and ground-truth answers, respectively. The accuracy of GPT-4o in answering questions related to the human's viewpoint in the image is only 27.5\%.}
    \label{fig:open}
\end{figure}

%% file: sections/02-Related-Work.tex
\section{Related Work}
\subsection{Large Multimodal Models}
Pre-trained on large scale of text corpus, Large Language Models (LLMs)~\cite{bert-2019,roberta-2019,gpt3-2020,zheng2023judging} can easily reach more than billions of parameters and show great capacity on natural language comprehension, text completion and generation, zero-shot transfer and in-context learning.
However, traditional LLMs usually take text as input and output, and lack the ability of understanding other modalities, like image, video or audio.

To tackle this issue, connecting multimodal encoder/decoder with LLMs, Large Multimodal Models (\LMMs) can integrate multiple data modalities and overcome the limitations of text-only LLMs.
\LMMs are utilized to address various different tasks: image-text understanding~\cite{liu2023llava,liu2023improvedllava,blip-2}, video-text understanding~\cite{lin2023video,video-chat,video-chatgpt}, and even multimodal generation~\cite{kosmos-2,kosmos-2.5,minigpt-5}.

In this paper, image-text understanding is our main focus, because it is already quite hard for \LMMs to understand the spatial relation in normal images, not to mention videos.

\subsection{Spatial relationship Benchmarks}
Though \LMMs have demonstrate remarkable performance in various benchmarks, they still have weaknesses on understanding spatial relationships, such as distinguish ``left'' from ``right'' between two objects presented in one image.

\citet{whatsup} curate a new What'sUP benchmark to quantify model performance on understanding spatial relationships.
The evaluation results show not only the pretraining corpus contains little spatial-related data to learn from, but also \LMMs perform limited on this spatial benchmark.
\citet{spatialvlm} identify the similar problem in 3D spatial relations.
By generating a large scale of spatial annotations, they increase the \LMMs spatial reasoning skill by a large margin.
Instead of increasing the scale of pre-training data, \citet{scaffolding} deal with spatial relationships from the model perspective.
They propose \textsc{Scaffold} prompting that overlays a dot matrix onto the image as visual information anchors,
which demonstrates better performance in spatial reasoning over GPT-4 Vision with the textual CoT prompting.

In this paper, we propose a new spatial reasoning benchmark Spatial-MM, covering a diverse spatial relationships more than What'sUP and containing less noisy or ambiguous annotations.

\subsection{Multi-hop Reasoning}
Chain of thought prompting~\cite{chain-of-thought} demonstrates remarkable multi-hop reasoning capability of LLMs by eliciting step-by-step reasoning paths. 
Least-to-most prompting~\cite{least-to-most} further shows the feasibility of conducting decomposition and multi-hop reasoning, which happens on the decoder side together with the answer prediction procedure.
Furthermore, \citet{II-MMR} improves multimodal multi-hop reasoning in VQA.
Using answer prediction-guided CoT, the II-MMR model finds a reasoning path to reach the answer.

Our new benchmark dataset Spatial-MM contains spatial multi-hop reasoning questions, covering at least two reasoning steps for each ground-truth reasoning path.
To the best of our knowledge, it is the first multimodal multi-hop benchmark that pays attention to the evaluation on the reasoning path.

%% file: sections/03-Spatial-MM-Benchmark.tex
\section{The Spatial-MM Benchmark}
\label{sec:spatial-mm}
We seek to study the spatial reasoning capability gap between humans and \LMMs. 
Based on the observation that existing benchmarks only partially investigate the spatial reasoning capabilities of \LMMs \cite{li2023seed, liu2023mmbench}, 
we introduce a novel benchmark, Spatial-MM, which includes two subsets: \datasetNameObj and Spatial-CoT. \datasetNameObj features multiple-choice questions that focus on the spatial relationships between one or two objects in an image, while Spatial-CoT offers open-ended multi-hop questions.

\input{figures/benchmark_patterns}
\subsection{\datasetNameObj}
\label{sec:spatial-mm-vqa}
\datasetNameObj is a carefully curated benchmark that contains 2,000 multiple-choice questions with the aim of assessing \LMMs' spatial reasoning of one or two objects in a given image. 
With natural images downloaded from the Internet, we carefully selected images to construct diverse challenging multiple-choice questions, including both yes/no questions and wh-type questions. 

The dataset is constructed with two rounds of annotations. 
In the first round of annotation, we divided the images among three annotators and tasked them with selecting one or two objects in each image and compose a question-answer (QA) pair including a spatial relationship. 
Annotators were provided with question templates with objects placeholders, which they could use or customize according to their preference.
In the second round of annotation, we released batches of 200 QA pairs with their corresponding images. 
Another 10 annotators were tasked with reviewing these QA pairs to verify whether they were correct or incorrect/ambiguous. 
Corrections were made based on their feedback. 

This dataset covers 36 of the most commonly used spatial relationships \cite{marchi2021cross}, including \emph{right, left, attached to, touching, back, bottom, ahead, forward, backward, down, facing towards, facing away, top, beneath, beside, side, behind, under, on, in, front, below, above, over, middle, between, inside, outside, bottom right, bottom left, top right, top left, corner, close to, next to, near}. 
Moreover, we used GPT-4o to categorise \datasetNameObj into visual patterns such as ``object localization'', ``orientation and direction'', ``viewpoints'' and ``positional and relational context'', which pose significant challenges for \LMMs. The prompt is listed in Appendix \ref{appendix:prompt}.
Examples of these patterns can be seen in Figure \ref{fig:patterns}.

\subsection{Spatial-CoT}
\label{sec:spatial-mm-multi-hop}
Chain of Thought (CoT)-style prompting has been demonstrated to significantly improve LLMs' reasoning capabilities. However, recent investigations on knowledge graph question answering and mathematical reasoning show that discrepancies exist between the answer and the corresponding reasoning paths through CoT prompting~\cite{nguyen2024direct,zhou2024don}. That is, while, by using CoT to producing reasoning paths, the LLM can produce the correct answer, the generated reasoning paths are not always correct. This disparity is a form of hallucination, and renders the LLMs' reasoning less trustworthy. 

To enable the study of faithfulness in \LMMs' spatial reasoning in the CoT style, we curate Spatial-CoT, a dataset of multi-hop question-answer pairs. 
The QA pairs are generated by prompting GPT-4o with a given image and a set of in-context examples.  
Images are sourced from Internet. 
In total, 800 multi-hop QA pairs were generated, and we filtered out 178 that did not include at least one of the 36 spatial relationships listed in Section \ref{sec:spatial-mm-vqa} above. 
We then employed human annotators to select reasonable and meaningful multi-hop QA pairs that require at least two reasoning steps to reach the final answer. 
An additional 312 QA pairs were manually discarded for lacking the complexity needed for multi-hop QA pairs. 
Ultimately, Spatial-CoT includes 310 spatial-aware multi-hop QA pairs. 
The prompt, including instructions for generating multi-hop QA, is provided in Section A of the appendix. 

\textbf{Spatial-aware reasoning paths.} In knowledge graph question answering, triplets are considered reasoning steps (hops)~\cite{nguyen2024direct}. Leveraging this idea, in VQA tasks, we utilize scene graphs, including object relations or attribute, as reasoning steps that lead to the answer. It is important to note that current VQA benchmarks (e.g.\ GQA~\cite{gqa}) with ground-truth scene graphs lack diverse perspective information and only incorporate spatial relationships from the camera's viewpoint. For Spatial-CoT, as no ground-truth paths (or scene graph) exist, to evaluate the spatial reasoning abilities of \LMMs, we generate the reasoning paths for each question. 
We carefully instruct GPT-4o with a set of in-context examples to generate the initial draft of the spatial-aware reasoning path. 
Subsequently, three annotators were tasked with evaluating (i.e.\ keeping, removing, adding, or modifying) the generated steps for each question. 
Additionally, the annotators were asked to tag the steps as $S$ for spatial hops such as ``person in front of car'' and $NS$ for non-spatial hops such as ``woman holding a fork'', respectively. In total, 67\% of reasoning steps are tagged as $S$ and the rest 33\% are tagged as $NS$, suggesting that many questions require reasoning of spatial relationships between key objects to be answered.

Each reasoning path includes two or more hops, with at least one being a spatial hop. Figure \ref{fig:reasoning_path} shows some examples of multi-hop QAs with their corresponding Reasoning path.
As shown in Table \ref{tab:spatial_hop}, among the questions, 34\% have a reasoning path with two hops, 39\% have three hops, and 27\% require four or more hops to answer.

\subsection{Human performance} 
We evaluated human performance using our benchmark to maintain the quality of annotations. 
We sampled 300 data points from \datasetNameObj and 100 samples from Spatial-CoT. 
Annotators were consulted to assess whether the correct option is clearly identifiable or if there's any potential for confusion. 
This human performance assessment yielded scores of 98\% and 99\% on \datasetNameObj and Spatial-CoT datasets, respectively.
\input{figures/reasoning_path}

%% file: figures/benchmark_patterns.tex
\begin{figure*}[!ht]
    \centering
    \resizebox{.90\textwidth}{!}{
    \includegraphics{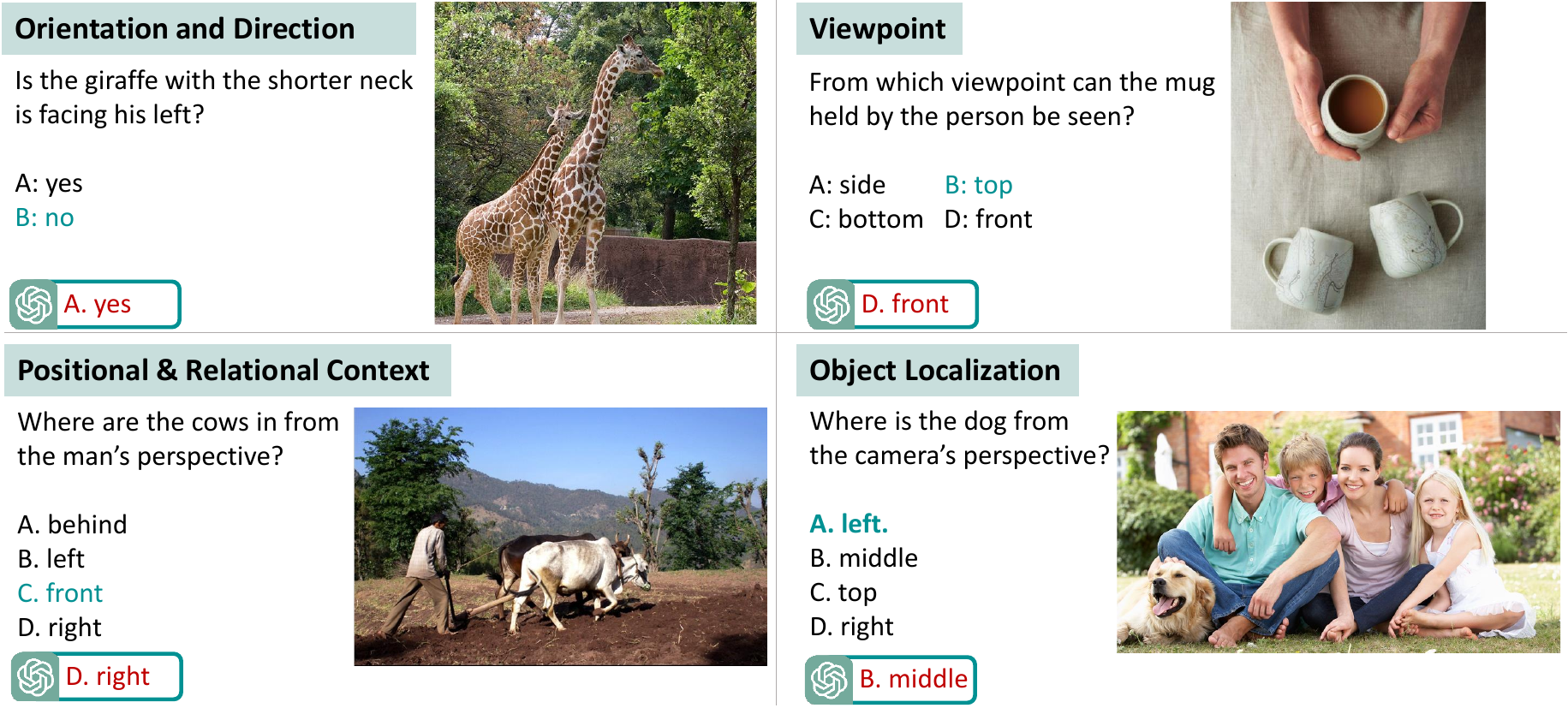}}
    \caption{VQA examples from our Spatial-MM that encompass a range of challenging visual patterns..}
    \label{fig:patterns}
\end{figure*}

%% file: figures/reasoning_path.tex
\begin{figure}[ht]
    \centering
    \resizebox{.4\textwidth}{!}{
    \includegraphics{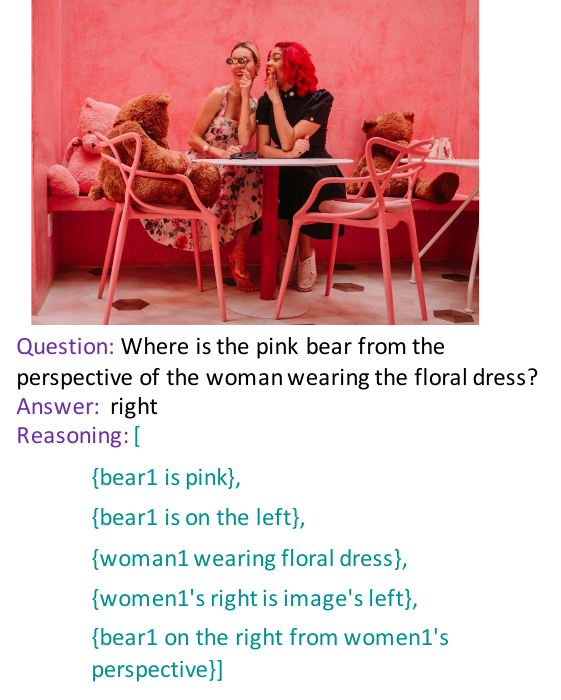}}
    \caption{An example of generated reasoning steps for a multi-hop question.}
    \label{fig:reasoning_path}
\end{figure}

%% file: sections/04-Methodology.tex
\section{Data Enrichment to Improve Spatial Reasoning}
\label{sec:methodology}
Intuitively, providing \LMMs with additional spatial-aware visual grounding data would help improve their spatial reasoning performance. 
To test this hypothesis, we describe two pipelines for generating different types of visual grounding data. The first pipeline consists of two stages: (I) key object extraction and (II) bounding box generation.
The second pipeline consists of three stages: (I) key object extraction, (II) spatial-aware caption generation related to the key objects, and (III) spatial-aware scene graph generation. 
Please refer to Figure \ref{fig:pieline} in Appendix \ref{appendix:pipeline} for detailed illustrations of our pipelines.
Furthermore, we describe Reasoning path generation for the multi-hop questions.

\noindent\textbf{Bounding box generation.}
The input of the pipeline is an image $I$ paired with a multiple-choice question $Q$. We first extract key objects, $[object_1, object_2,..]$ from the given question. Next, we prompt GPT-4o to provide the bounding boxes of the key objects in the image. Each bounding box is specifically represented as a tuple $[x_{min}, y_{min}, x_{max}, y_{max}]$, where $x_{min}$ and $y_{min}$ are coordinates of the top-left corner and $x_{max}$ and $y_{max}$ are coordinates of the bottom-right corner. 
We present the prompt for bounding box generation in Appendix \ref{appendix:prompt}, and examples of synthesized bounding boxes for the key objects along with the ground-truth bounding boxes in Appendix \ref{appendix:boundingbox}.

\noindent\textbf{Scene graph generation.} Similar to the first stage of bounding box generation, we extract key objects from the given question. We then prompt GPT-4o to generate a spatial-aware caption by considering the location, direction, orientation, and spatial relations of the key objects in the image. Finally, given the question, spatial-aware caption, key objects, and image, GPT-4o is prompted to extract spatial relation, orientation, and direction of the key objects in the image. The prompt for scene graph generation is listed in the Appendix \ref{appendix:prompt}.

\noindent\textbf{Reasoning path generation.}
Previous research on evaluating \LMMs has solely focused on final answer accuracy, neglecting the correctness of the generated reasoning paths \cite{lei2024scaffolding,II-MMR}. In this work, we delve deeper into the spatial reasoning path of \LMMs in multi-hop VQA by utilizing the ground-truth reasoning steps we generated for multi-hop questions in Section \ref{sec:spatial-mm-multi-hop}. 
To address the challenge of evaluating unstructured reasoning steps, we carefully design a prompting strategy to instruct \LMMs to output the reasoning steps in a scene graph format. This enables us to  validate against ground-truth reasoning steps. Figure \ref{fig:reasoning_path} shows an example of generated reasoning paths for an image. 

%% file: sections/05-Experiment.tex
\section{Experiments}
\input{tables/bbox_sg_vqa}

\subsection{Dataset}
We use the following datasets in our experiments:

\noindent$\bullet$ \textbf{Spatial-MM} contains \textbf{\datasetNameObj} described in \S \ref{sec:spatial-mm-vqa} and \textbf{\datasetNameCoT} described in \S \ref{sec:spatial-mm-multi-hop}.

\noindent$\bullet$ \textbf{GQA-spatial}~\cite{whatsup} is sourced from GQA~\cite{gqa}, where each image is paired with two caption options, which contain opposite spatial relationships, i.e.\ \textit{prepositions}.
Depending on the number of \textit{object(s)} identified in the image, GQA-spatial contains 1-object and 2-object caption options. 
One instance from GQA-spatial is shown in the ``standard'' row of Figure \ref{fig:caption_perturbation}: one image and two captions make use of opposite \textit{preposition}s: \textit{front} and \textit{behind}.\footnote{There are three pairs of opposite spatial relationships in GQA-spatial: \{\textit{left, right}\}, \{\textit{front, behind}\} and \{\textit{top, bottom}\}.}

\subsection{\LMMs}
We evaluate the spatial reasoning capability of the following four \LMMs: \textbf{LLaVA-1.5-7B}~\cite{llava-v1.5}, \textbf{GPT-4 Vision}~\cite{gpt-4v}, \textbf{GPT-4o}~\cite{gpt-4o}, \textbf{Gemini 1.5 Pro}~\cite{gemini}, and \textbf{MiniGPT-v2 (7B)}~\cite{MiniGPT-v2}.
Please refer to Appendix \ref{appendix:LMMs} for details.

\subsection{Evaluation Metrics}
To evaluate model performance on final answer prediction, we adopt the standard Accuracy metric \citet{gqa}. 
For evaluating \datasetNameCoT, the exact-match metric is insufficient for measuring performance fully due to the open-ended nature of the answers. \LMMs frequently produce paraphrases or alternative expressions that convey the same underlying meaning.
Therefore, we initially verify if the predicted answer exactly matches the gold answer. If it does not, following \cite{lyu2024automatic, wang2023chatgpt}, we use GPT-4o as a judge to determine whether the predicted answer holds the same semantic meaning as the gold answer.

\subsection{The Effect of Bounding Boxes and Scene Graphs}
For both \datasetNameObj and GQA-spatial benchmarks, the input is an image paired with a multiple-choice question. 
Table \ref{tab:bbox_sg_vqa} presents the results on \datasetNameObj and GQA-spatial, comparing \LMMs' performance on images only and with additional information (i.e.\ bounding boxes and scene graphs). A number of important observations can be made from the table. (I) It shows that the additional bounding box information serves as visual anchors that enhance \LMMs' spatial reasoning performance for \LMMs. When given the synthesize bounding boxes information on GQA dataset, the \LMMs improve their answer accuracy significantly, with the average increase over ``stan'' of 33 points. However, spatial related questions require a profound semantic understanding of object relationships within the scene. By incorporating spatial-aware scene graphs of key objects, the performance of \LMMs improves significantly on both datasets. Specifically, by including the synthesized SG in the prompt, GPT-4 vision's accuracy increases by 9.28\% and 8.92\% for one-object and two-object questions on the \datasetNameObj dataset. (II) Surprisingly, the \LMMs perform better with synthesized bounding boxes than with ground-truth bounding boxes. This may be attributed to the high quality and diversity of synthesized bounding boxes generated by GPT-4 vision. (III) On \datasetNameObj, scene graphs further enhance model performance, while on GQA-spatial, bounding box information seems to be more useful. 
We leave a deeper analysis of these two observations to future work. 

\noindent\fbox{
\parbox{.95\columnwidth}{\noindent\textbf{Finding 1:} Bounding boxes and scene graphs enhance \LMMs’s ability to excel in visual reasoning tasks. Bounding boxes are more effective in one-object questions, while scene graphs are more helpful in two-object ones.}
}
\input{tables/perspective}

\subsection{The Effect of Human and Camera Perspectives}
Questions from previous benchmarks~\cite{whatsup} are often evaluated from the camera's perspective, i.e., \textbf{outside} the image.
How would \LMMs behave if the question is asked from other angles, such as from the human perspective in the image, i.e., \textbf{inside} the image, is under-explored.
This problem motivates us to generate different questions with distinct prompts ``from camera's perspective'', ``from human's perspective'' respectively, from original questions, and further understand the spatial reasoning capabilities of \LMMs.

In Table \ref{tab:perspective}, we show different model performance from either human's or camera's perspective on \datasetNameObj and \datasetNameCoT.
Though all these \LMMs receive explicit prompt ``from human/camera's perspective'', they all show significant performance drop on human's perspective, when compared to the camera's perspective.
Especially, while GPT-4o outperforms all other models from camera's perspective, it is still not able to understand from human's perspective and show the maximum drop on \datasetNameCoT of more than 30 points.

\noindent\fbox{
\parbox{.95\columnwidth}{\noindent\textbf{Finding 2:} \LMMs excel at understanding the scene from the camera's perspective. However, their performance declines significantly when questions are posed from the human perspective within the image.}
}

\input{tables/hop_reasoning}

\subsection{Analysis based on Complex Multi-hop VQAs}
\textbf{The effect of the number of hops to the answer.} 
Table~\ref{tab:spatial_hop} compares chain of thought prompting with other approaches on the \datasetNameCoT dataset. 
As can be seen, the conventional CoT prompting \cite{wei2022chain} may not be as effective for complex VQA tasks as it is for NLP tasks. Indeed, we can observe that even the strong \LMMs such as GPT-4o, Gemini-Pro and LLaVA-1.5 perform worse when employing CoT reasoning, compared to standard prompting. Therefore, the rationales produced by the conventional CoT may not align well with the reasoning path needed to arrive at the answer. 

On the contrary, including visual grounded information such as bounding boxes information is effective when the number of reasoning hops are $\le 3$.
In particular, Table \ref{tab:spatial_hop} highlights the effectiveness of scene graphs in answering multi-hop spatial questions across all the models. Multi-hop questions demand a deep semantic understanding of object attributes and relationships within the scene. Therefore, scene graphs are essential for enhancing LLMs' semantic visual understanding. 
Moreover, across different models and various experiment settings in Table \ref{tab:spatial_hop}, it can be seen that with the increase in the number of the hops, accuracy of the final answer drops. For instance, the average accuracy of questions with $\ge 4$ hops is 12\% and 7.5\% less than that for questions with 2 and 3 hops, respectively. 

\input{tables/reasoning_path}
\input{tables/eval_reasoning_path}
 \paragraph{Correctness of the reasoning path.}
We ground \LMMs’ reasoning steps into a scene graph format and verify whether they form a valid path.
Given the question and predicted answer, we carefully design prompts to instruct \LMMs to output the reasoning steps in a scene graph format, i.e., including object relations and/or attributes. This enables us to validate the \LMMs' reasoning steps against ground-truth reasoning paths (Section \ref{sec:methodology}). To evaluate the reasoning path in multi-hop questions, computing the \emph{semantic match} for all the generated steps in the path.
We further analyze the types of reasoning steps required in multi-hop questions. 
 
Table \ref{tab:eval_reasoning} presents the results of analyzing the reasoning paths for questions that \LMMs fail to answer correctly by $stan$ prompting. The incorrect reasoning path could occur due to an incorrect spatial step, an incorrect non-spatial step, or a combination of both. On average, only 1\% of questions had the correct reasoning path but an incorrect final answer. Moreover, 91\% of questions included at least one incorrect spatial reasoning step, while only 9.75\% contained at least one incorrect Non-spatial reasoning step. Please note that an incorrect reasoning step occurs when a reasoning step semantically does not match the ground truth reasoning steps or when a reasoning step from the ground truth path is missing in the generated path. 

The average scores for all Spatial and Non-spatial reasoning steps in \datasetNameCoT questions are listed in Table \ref{tab:reasoning_path}, which shows that the average gap between F1 scores for \textit{Spatial} and \textit{Non-spatial} reasoning steps is significant by 21 point on $stan$ and 19 points on \textit{stan+SG$_{\text{synth}}$} prompting. 

\noindent\fbox{
\parbox{.95\columnwidth}{\textbf{Finding 3:} Chain of thought prompt is not effective for multi-hop spatial reasoning. 
}}


\subsection{Analysis of Perturbations on GQA-spatial}
\input{figures/caption_perturbation}
Using GQA-spatial, we do a comprehensive understanding of the \LMMs' robustness when facing spatial-related questions by applying \textbf{five perturbation settings} on the caption options as follows:
\begin{itemize}[leftmargin=*]
\setlength\itemsep{0em}
    \item {none}: adding one extra option ``None of above''; the ``Standard+None'' row in Figure \ref{fig:caption_perturbation}
    \item {rel\_neg}: adding ``not'' to the correct option;  the ``Relation Changed (Negation)'' row in Figure \ref{fig:caption_perturbation}
    \item {rel\_swap}: Swapping the \textit{key objects} in options;  ``Relation Changed (Swapping)'' row in Figure \ref{fig:caption_perturbation}
    \item {obj\_change}: replacing one of the key objects for both options (A \& B); the ``Object Changed (Two options)'' row in Figure \ref{fig:caption_perturbation}
    \item {obj\_change\_a}: replacing one of the key objects in the correct option (A) with a none-existing object in the image; the ``Object Changed (option A)'' row in Figure \ref{fig:caption_perturbation}
\end{itemize}

For the Standard setting, the model input is an image paired with two caption options that differ only by the \textit{preposition} they contain, from which \LMMs should be able to select the caption option with the correct \textit{preposition}.
For perturbation settings, the image remains the same but caption options are changed with different difficulty levels.
Intuitively speaking, we posit that all perturbation settings are more difficult than the Standard setting, as they add noise (e.g., none) and create obstacles (e.g., negation) for \LMMs.
Results are show in Table~\ref{tab:perturbstion-gqa}, and analysis of each perturbation are discussed in detail below:

\input{tables/perturbation-gqa}


\noindent\textbf{rel\_neg.}
Adding \emph{not} to the answer option A increases the accuracy (7\%) for 1-object only, meanwhile decreasing the accuracy (6\%) for 2-objects.
Negating the relations in general decreases model performance. Adding bounding boxes makes it even worse. However, adding scene graphs is helpful, especially with Gemini.

\noindent\textbf{rel\_swap.}
Analysis is conducted with swapping objects of answer option A and adding the ``None'' option. From 288 two\_obj questions, Gemini only has 8 correct answers ( option E), 5 incorrect answers(option A), and 275 incorrect answers(option B), while 95\% of generated answers are option B.

LLaVA shows similar results: from 291 2-object questions, LLaVA only has 4 correct answers (option E), 3 incorrect answers(option A), and 284 incorrect answers(option B), while 97.59\% of generated answers are option B.


\noindent\textbf{obj\_change.}
When the same object in both options A and B are swapped with an object that does not exist in the image (row ``obj\_change+none'' in Table~\ref{tab:perturbstion-gqa}),  both Gemini and LLaVA show the highest accuracy scores in both one\_obj and two\_obj ((e.g. 91.87\% and 69\% respectively for Gemini). When only changing the object in the correct option (A), both models' performance decreases significantly. For instance, Gemini's one\_obj performance decrease from 91.87\% to 46\% and two\_obj performance from 69.10\% to 30\%. 


\noindent\fbox{
\parbox{.95\columnwidth}{\textbf{Finding 4:}
\LMMs are usually good at the object detection task (recognizing objects present in the image), but struggle with spatial reasoning (distinguishing ``left'' from ``right'').
}}

We also conducted an additional analysis of spatial propositions of \emph{front-behind}, \emph{left-right} and \emph{top-bottom} in Appendix~\ref{sec:preposition}. Interestingly, Gemini, GPT-4 vision and LLaVA show different performance characteristics across the propositions. 




%% file: tables/bbox_sg_vqa.tex
\begin{table*}
    \centering
    \small
    \resizebox{0.7\linewidth}{!}{
    \begin{tabular}{l|lll|lll|lll|lll}
    \hline
 & \multicolumn{3}{c|}{GPT-4 vision} & \multicolumn{3}{c|}{Gemini} & \multicolumn{3}{c}{LLaVA-1.5} &
 \multicolumn{3}{c}{MiniGPT-v2} \\
  Prompt  & 1\_obj   & 2\_obj & all& 1\_obj   & 2\_obj & all & 1\_obj   & 2\_obj & all& 1\_obj   & 2\_obj & all\\ 
\hline
 \multicolumn{13}{c}{\cellcolor{gray!30}\datasetNameObj}  \\
\hline
 stan &50.80& 52.85& 52.18 &46.63&41.53 & 43.18 &51.59 &43.58& 46.18&45.73&35.83&39.99\\
 stan+bbox$_{\text{synth}}$& \textbf{71.34} &60.21 & \textbf{63.82} & 50.81&45.55 & 47.26 &\textbf{56.97} &44.62& 48.63&\textbf{53.61}&40.87&	\textbf{44.93}\\
 stan+SG$_{\text{synth}}$& 60.08& \textbf{61.77}& 61.22 &\textbf{53.42} &\textbf{55.45} & \textbf{54.79} & 51.60&\textbf{50.79}& \textbf{51.05}& 49.44	& \textbf{41.39}	&43.95\\ \hline
  \multicolumn{13}{c}{\cellcolor{gray!30}GQA-spatial}  \\

\hline
 stan&66.46&15.81&56.30&19.40&16.35&18.79&27.16&15.12&24.75&22.14&8.63&19.43\\
 stan+bbox$_{\text{gt}}$&79.69&40.89&71.91&33.74&37.84&33.76&68.19&42.96&63.13 & 33.45&	21.69&	31.09 	 \\    stan+bbox$_{\text{synth}}$&\textbf{85.86}&64.95&\textbf{81.67}&\textbf{42.11}&48.00&43.29&\textbf{80.86}&\textbf{50.52}&\textbf{74.78}& 50.26	&39.99&	48.20\\     stan+SG$_{\text{gt}}$&71.60&\textbf{70.69}&71.42&41.36&\textbf{55.21}&\textbf{44.14}&55.40&36.90&51.69&39.67&	31.69	&38.07\\
 stan+SG$_{\text{synth}}$&64.90&55.40&62.99&24.77&49.25&29.68&36.99&46.55&38.91&51.77&	33.9&	48.19\\\hline
\end{tabular}}
\caption{The accuracy of models in the Standard (stan) setting, augmented with synthesized bounding boxes (bbox$_{\text{synth}}$), scene graphs (SG$_{\text{synth}}$), ground truth bounding boxes (bbox$_{\text{gt}}$) and ground truth scene graphs(SG$_{\text{synth}}$ ).
 The synthetic bounding boxes and scene graphs are produced by GPT-4o. \textbf{Bold}: Best results within each dataset.
}
\label{tab:bbox_sg_vqa}
\end{table*}

%% file: tables/perspective.tex
\begin{table}[t]
\resizebox{.9\linewidth}{!}{
\begin{tabular}{l|c|c|c}
\hline
 & Human's pers. & Camera’s pers. & Overall \\ \hline
\multicolumn{4}{c}{\cellcolor{gray!30}\datasetNameObj} \\ \hline
Distribution & 31\%& 69\%& 100\% \\ \hline
GPT-4o   & 43.75  & 73.6   & 64.89   \\
GPT-4 vision & 42.56&  54.81 & 51.23  \\
LLaVA  &  37.70 & 46.04 & 43.61   \\
Gmini  & 32.92  & 44.07 & 40.82 \\
MiniGPT-v2& 31.67 & 40.75 & 37.94 \\\hline
 \multicolumn{4}{c}{\cellcolor{gray!30}\datasetNameCoT} \\ \hline
Distribution & 46\%& 54\%& 100\% \\ \hline
GPT-4o   & 27.50  & 63.25& 46.81   \\
GPT-4 vision  & 33.05  &50.14 &  42.28  \\
LLaVA  &  22.91 & 40.47&  32.39  \\
Gmini  & 25.81  &46.29 &  36.87  \\
MiniGPT-v2  &  22.08 & 39.77 & 31.63 \\\hline

\end{tabular}}
\caption{Results of human and camera perspectives.}
\label{tab:perspective}
\end{table}

%% file: tables/hop_reasoning.tex

\begin{table}[]
\resizebox{0.99\linewidth}{!}{
\begin{tabular}{ll|cccc}
\hline
\multicolumn{2}{l}{Num of hops} & 2-hop & 3-hop & $\geq 4$-hop & All \\ \hline
\multicolumn{2}{l}{Hop Distribution} & 34\% & 39\% & 27\%  & 100\% \\ \hline
\multirow{4}{*}{GPT-4o}& stan & 76.47 & 60.00 & 42.86 & 60.78\\
& stan+\small{COT} &76.47 & 55.54 & 46.89 & 56.91\\
&stan+bbox$_{\text{synth}}$&\textbf{82.35} & 55.0 & 35.71 & 58.82 \\
& stan+SG$_{\text{synth}}$&73.33 & \textbf{75.0} & 35.71 & 63.27\\\hline
\multirow{4}{*}{GPT-4v} &stan& 58.82 & 55.0 & 50.0 & 54.9 \\
& stan+\small{COT} & 50.0 & 60.0 & \textbf{60.0} & 56.86\\
&stan+bbox$_{\text{synth}}$&56.25 & 50.0 & 50.0 & 52.0\\
& stan+SG$_{\text{synth}}$ & 81.25 & \textbf{75.0} & 35.71 & \textbf{66.0} \\\hline
\multirow{4}{*}{LLaVA} &stan& 47.06 & 50.0 & 50.0 & 49.02\\
& stan+\small{COT} &35.29 & 55.0 & 21.43 & 39.22 \\
&stan+bbox$_{\text{synth}}$&47.09 & 50.0 & 35.71 & 46.14\\
& stan+SG$_{\text{synth}}$ & 52.94 & 60.0 & 50.0 & 54.9\\\hline
\multirow{4}{*}{Gmini}&stan&52.94 & 50.0 & 42.86 & 49.02\\ 
& stan+\small{COT} &33.33 & 29.41 & 36.84 & 33.33\\
&stan+bbox$_{\text{synth}}$&56.25 & 30.0 & 28.57 & 38.0\\
& stan+SG$_{\text{synth}}$ &68.75 & 55.0 & 57.14 & 60.0 \\\hline
\end{tabular}}
\caption{We evaluated the accuracy of \LMMs on different numbers of hops on our challenging multi-hop benchmark, \datasetNameCoT.}
\label{tab:spatial_hop}
\end{table}

%% file: tables/reasoning_path.tex
\begin{table}[]
\resizebox{0.99\linewidth}{!}{
\begin{tabular}{ll|lll|lll}
\hline
\multicolumn{2}{l|}{Reasoning type} & \multicolumn{3}{c|}{Spatial} & \multicolumn{3}{c}{Non-spatial} \\ 
\multicolumn{2}{l|}{Distibution of steps} & \multicolumn{3}{c|}{67\%} & \multicolumn{3}{c}{33\%} \\ \hline
Models & Prompting & P & R & F1 & P & R & F1 \\ \hline
\multirow{2}{*}{GPT-4o} & stan &79.28 & 63.77& 70.28& 90.48& 95.02& 92.68 \\
 & stan+SG$_{\text{synth}}$ &80.81 &68.38 &74.07 & 91.43& 96.00 & 93.66 \\\hline
\multirow{2}{*}{Gemini} &stan&66.99 & 69.70&68.32& 86.27 & 89.80 & 88.01 \\
 & stan+SG$_{\text{synth}}$ & 80.21& 64.38&71.43 &90.10&92.86 & 91.46 \\\hline
\multirow{2}{*}{LLaVA} & stan & 66.02&71.58 & 68.96& 85.62 & 92.71 & 89.02 \\
 & stan+SG$_{\text{synth}}$ &77.23&67.83&72.22&82.41 &93.68 & 87.68 \\ \hline
\end{tabular}}
\caption{Evaluation of Spatial and Non-spatial reasoning steps on our \datasetNameCoT benchmark.}
\label{tab:reasoning_path}
\end{table}

%% file: tables/eval_reasoning_path.tex
\begin{table}[]
\resizebox{0.99\linewidth}{!}{
\begin{tabular}{l|c|cc}
\hline
Reasoning Path & Correct & \multicolumn{2}{c}{Incorrect}             \\ \hline
Reasoning Type & None    & \multicolumn{1}{c}{Spatial} & Non-spatal \\ \hline
GPT-4o         & 0.53    & 92.55   & 6.91       \\
GPT-4 vision   & 1.18    & 91.76   & 8.24       \\
LLaVA          & 1.99    & 89.40   & 12.58      \\
Gmini          & 1.32    & 90.73   & 11.26      \\ \hline
Average (\%)       & 1.25    & 91.11   & 9.75       \\ \hline
\end{tabular}}
\caption{Evaluation of reasoning path on multi-hop questions which \LMMs fail to answer correctly.}
\label{tab:eval_reasoning}
\vspace{-2mm}
\end{table}

%% file: figures/caption_perturbation.tex
\begin{figure}[t]
    \centering
    \resizebox{.49\textwidth}{!}{
    \includegraphics{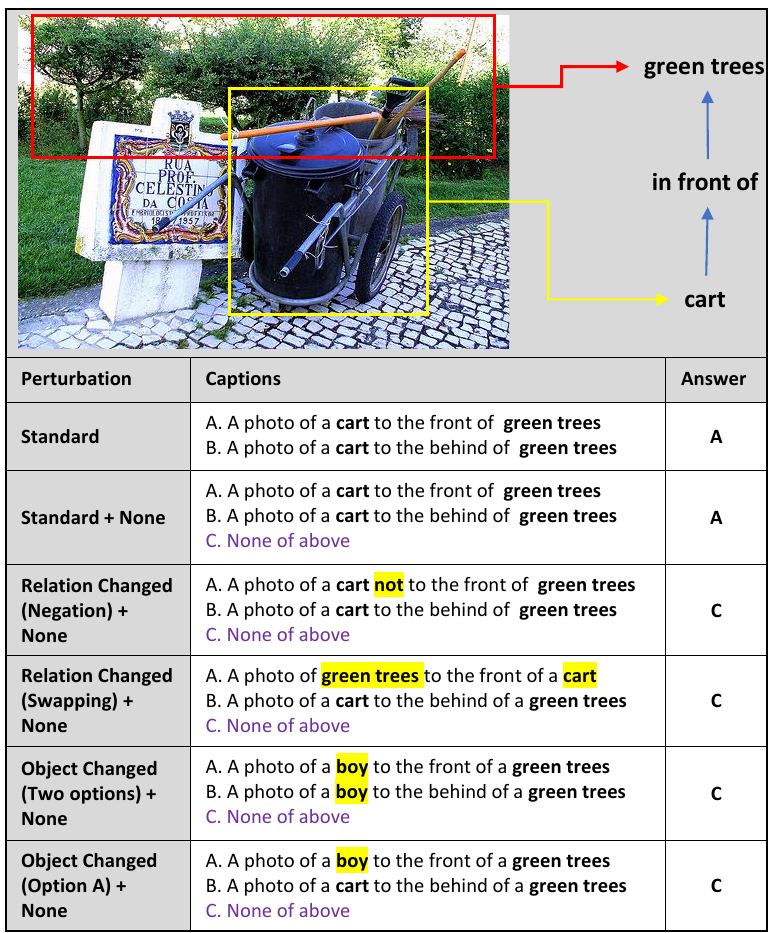}}
    \caption{Examples of caption perturbation.}
    \label{fig:caption_perturbation}
\end{figure}

%% file: tables/perturbation-gqa.tex
\begin{table}
    \centering
    \resizebox{1\linewidth}{!}{
    \begin{tabular}{l|ll|ll}
    \hline
Model  & \multicolumn{2}{c|}{Gemini-Pro} & \multicolumn{2}{c}{LLaVA-1.5} \\\hline
Data (GQA)   & one\_obj   & two\_obj  & one\_obj  & two\_obj  \\
\hline
 stan&19.40&16.35&27.16&15.12\\
 stan+none&19.14&22.01&33.02&15.12\\
 stan+none+bbox$_{\text{gt}}$&30.98&39.24&37.16&15.46\\ \hline
 rel\_neg+none&1.75&2.176&41.38&8.93\\
 rel\_neg+none+bbox$_{\text{gt}}$&0.43&0.35&16.98&2.41\\
 rel\_neg+none+SG&33.19&50.69&38.19&6.87\\
 rel\_swap+none&\_&2.78&\_&1.37\\
 rel\_swap+none+bbox$_{\text{gt}}$&\_&2.07&\_&0.00\\  
 rel\_swap+none+SG$_{\text{gt}}$&\_&64.58&\_&0.00\\\hline
 obj\_change+none&91.87&69.10&90.69 &54.98\\
 obj\_change\_a+none&45.76&30.21&\emph61.55 &52.58\\
\hline
\end{tabular}}
\caption{Results of different types of analyses on GQA-spatial dataset. }
\label{tab:perturbstion-gqa}
\vspace{-8pt}
\end{table}

%% file: sections/06-Conclusion.tex
\section{Conclusion}
Motivated by our observation that \LMMs are not able to distinguish spatial relationships, in this paper, we propose a new benchmark to evaluate the spatial reasoning capabilities of \LMMs.
Our benchmark consists of two subsets: \datasetNameObj containing 2,000 multiple-choice questions to evaluate the spatial reasoning capabilities of one or two objects in a given image, \datasetNameCoT containing 310 multi-hop questions to evaluate the spatial related CoT and reasoning paths.
We conduct comprehensive experiments on our benchmark and GQA-spatial.
Experimental results show the deficiencies in spatial reasoning of current \LMMs, including GPT-4o, Gemini and LLaVA-v1.5.
We also find that bounding boxes and scene graphs are helpful in some cases to improve the \LMMs' prediction quality.

%% file: sections/limitations.tex
\section{Limitation}
To set up the \datasetNameCoT benchmark, we implement a two-step data collection process: initially using GPT-4 to generate multi-hop VQAs, followed by manual filtering and modification. 
Though this process can to some extend accelerate the overall annotation approach as human annotators don't need to write multi-hop VQAs from the scratch, it is still limited and hard to create and generate large scale of labelled data.
Same limitations applies to the generation of the ground-truth reasoning paths.

%% file: sections/appendix.tex
\label{appendix:prompt}
\input{figures/appendix_prompt1}
\input{figures/appendix_bbx-generation_prompt}

\begin{figure*}[h!]
    \centering
    \resizebox{.99\textwidth}{!}{
    \includegraphics{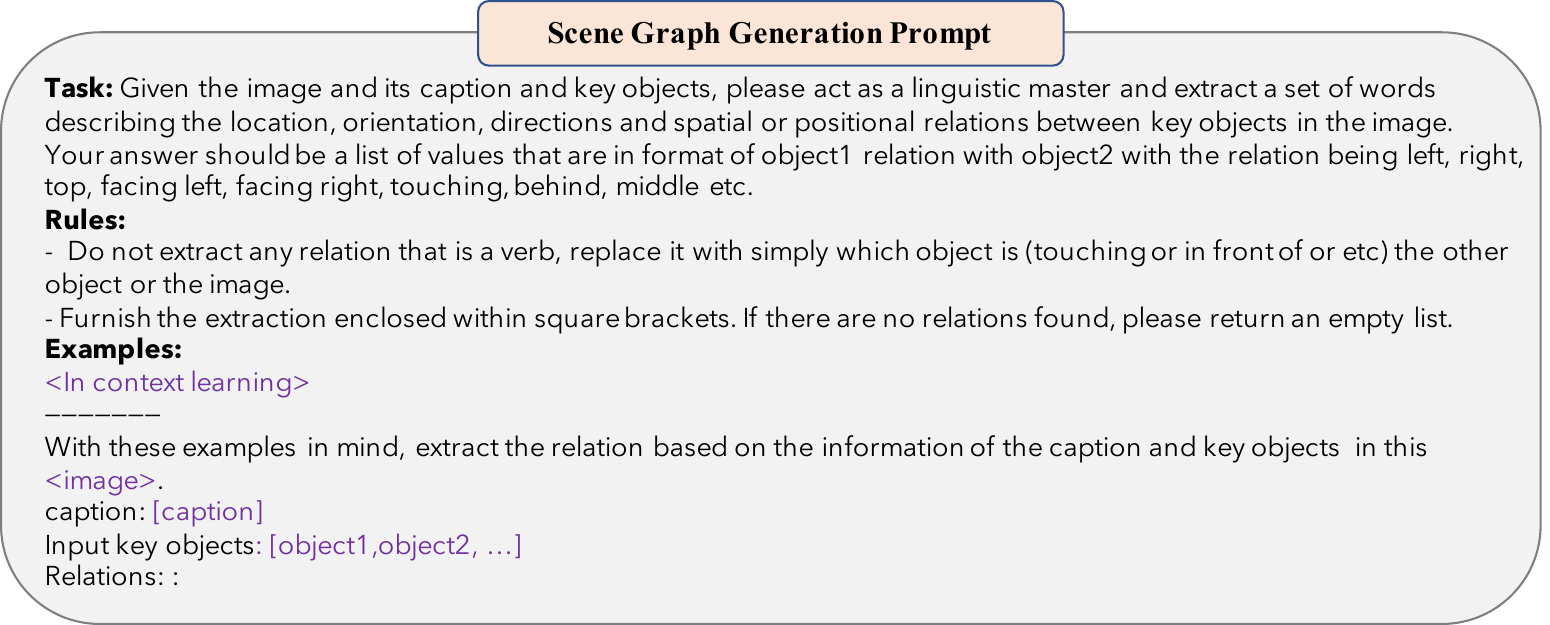}}
\end{figure*}

\begin{figure*}[h!]
    \centering
    \resizebox{.99\textwidth}{!}{
    \includegraphics{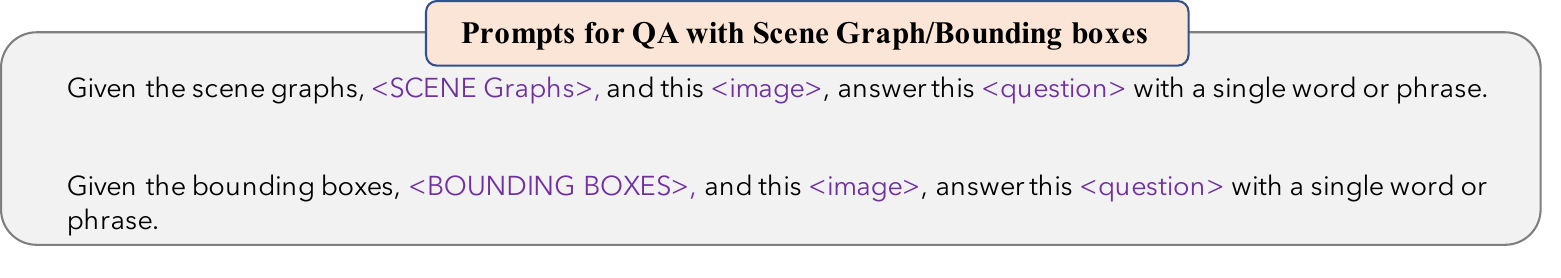}}
\end{figure*}

\begin{figure*}[h!]
    \centering
    \resizebox{.99\textwidth}{!}{
    \includegraphics{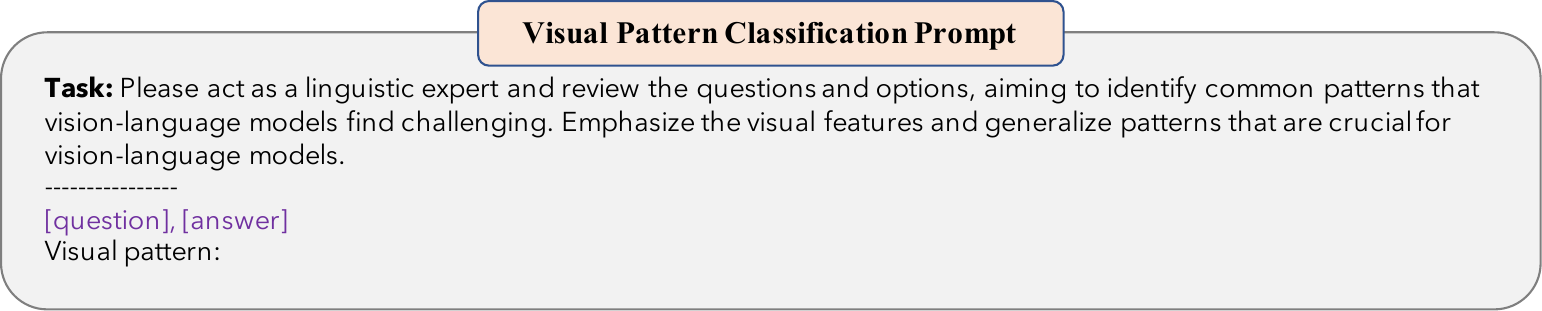}}
\end{figure*}

\section{\LMMs}
\label{appendix:LMMs}
In this paper, we evaluate the spatial reasoning capability of the following four \LMMs:
\noindent$\bullet$ \textbf{LLaVA-1.5-7B}~\cite{llava-v1.5} is an open-source multimodal instruction-tuned model that achieves state-of-the-art performance on 11 benchmarks, with just simple modifications to the original LLaVA~\cite{llava}. 

\noindent$\bullet$ \textbf{GPT-4 Vision}~\cite{gpt-4v} is a closed-source \LMMs, which is an advanced capability of OpenAI's GPT-4 model.
It is powerful to understand visual inputs such as images, and generate textual responses based on the content of images.

\noindent$\bullet$ \textbf{GPT-4o}~\cite{gpt-4o}, also known as GPT-4 Omni, is the latest and most advanced AI model.
Improved upon GPT-4, it is able to handle multimodal inputs and outputs, including text and images, with the enhancement of multimodal capabilities and speedup over previous models.

\noindent$\bullet$ \textbf{Gemini 1.5 Pro}~\cite{gemini} is a closed-source \LMMs from Google Deepmind.
Gemini 1.5 Pro is designed to be highly capable and general, excelling across various visual tasks with state-of-the-art performance. 

\noindent$\bullet$ \textbf{MiniGPT-v2 (7B)}~\cite{MiniGPT-v2} is an improved version of MiniGPT, an open-source large multimodal model that enhances the ability to comprehend and describe visual information by combining a lightweight architecture with strong visual-language reasoning capabilities.

\section{Spatial-MM Benchmark Samples}
The Figure \ref{fig:examples} shows sampled instances from Spatial-MM benchmark.
\input{figures/examples_fig}

\section{Pipeline}
Figure \ref{fig:pieline} illustrates our pipelines for data enrichment.
\label{appendix:pipeline}
\input{figures/piplibe}

\section{Bounding Box Example}
\label{appendix:boundingbox}
\input{figures/bbx_visulization}
Figure \ref{fig:bbx} shows examples of synthesized bounding boxes for the key objects along with the ground-truth bounding boxes.
\section{Spatial Relationships Types}
Figure \ref{fig:rel_distribution} demonstrates the distribution of spatial relationships in Spatial-MM.
\input{figures/spatial_rel_distribution}

\section{Preposition analysis}\label{sec:preposition}
In Table \ref{tab:preposition-analyses}, we present the model performance on different spatial relationships.

\textbf{Gemini-Pro.} For the \textit{left/right} relations, including bounding boxes increases the accuracy(16-18\%). However, by including SG, the accuracy increases significantly(26-37\%)\\
For the \textit{bottom/top} relations,
while adding bounding boxes of the objects increases the accuracy slightly(one-obj only)(5\%), adding SG increases the accuracy for one-obj questions by 15\%.\\
For the \textit{behind/front} relations, additional bounding box information does not change the accuracy or make it worse. however, including SG increases the accuracy significantly(for two-obj questions)(42\%).

\textbf{GPT-4-Vision.}
For the \textit{left/right} relations, adding bounding boxes and SG increases the accuracy by 29\% and 65\%, respectively.
Regarding \textit{bottom/top} relations, while adding bounding boxes does not change the accuracy, adding SG increases the accuracy.
For the \textit{behind/front}, while including the bounding box, drops the accuracy by 13\%, adding SG increases the accuracy by 28\%.

\noindent\fbox{
\parbox{.95\columnwidth}{\textbf{Additional Finding}: The three LMMs exhibit different performance characteristics across the three spatial propositions. 
}}

\input{tables/preposition_analyses}

\section{Prompts}

%% file: figures/appendix_prompt1.tex
\begin{figure*}[h]
    \centering
    {
    \includegraphics[width=.99\textwidth]{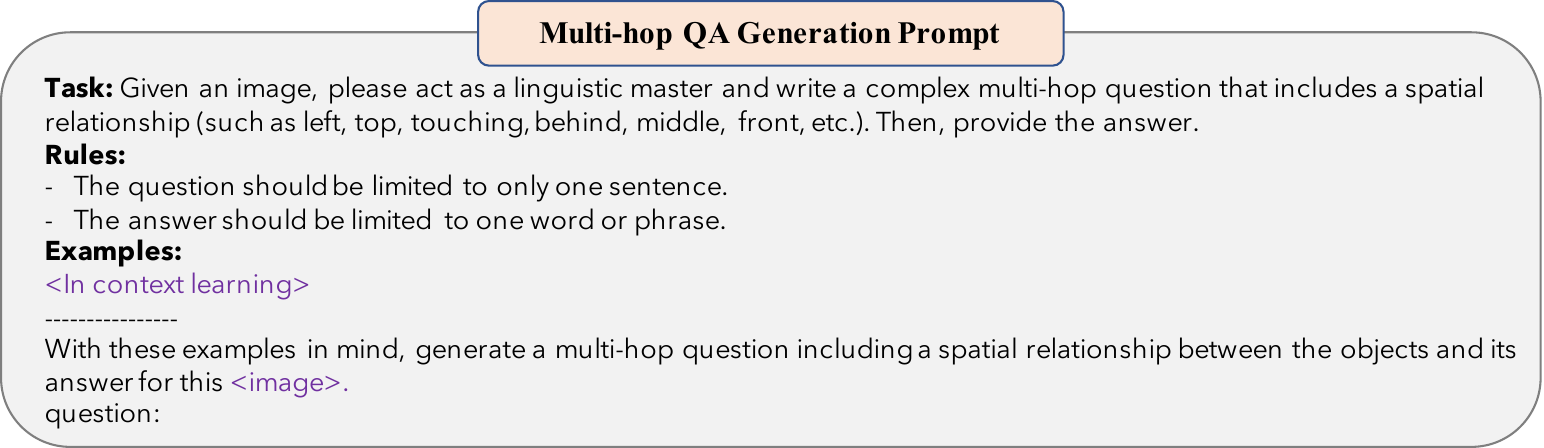}}
    \label{fig:multihop_qa}
\end{figure*}

%% file: figures/appendix_bbx-generation_prompt.tex
\begin{figure*}[h]
    \centering
    \resizebox{.99\textwidth}{!}{
    \includegraphics{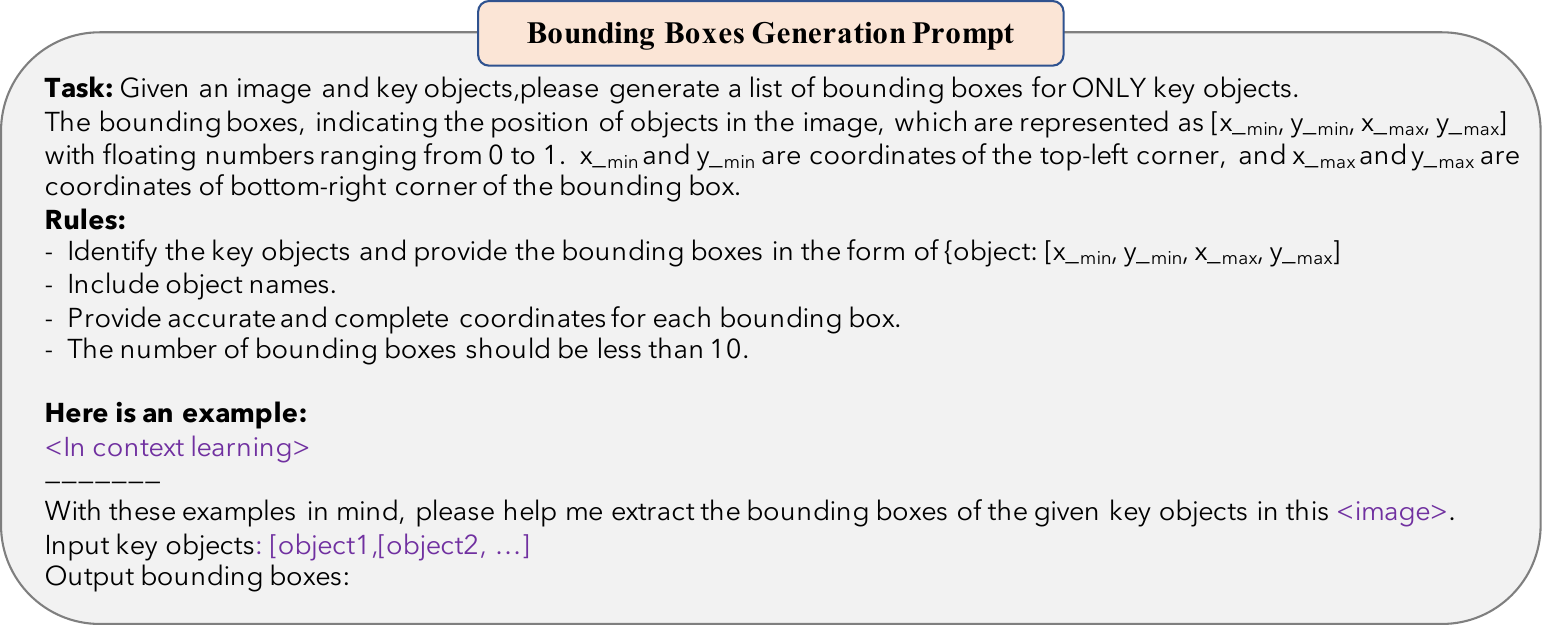}}
    \label{fig:bbx_prompt}
\end{figure*}

%% file: figures/examples_fig.tex
\begin{figure*}[h!]
    \centering{
    \includegraphics[width=1.0\textwidth]{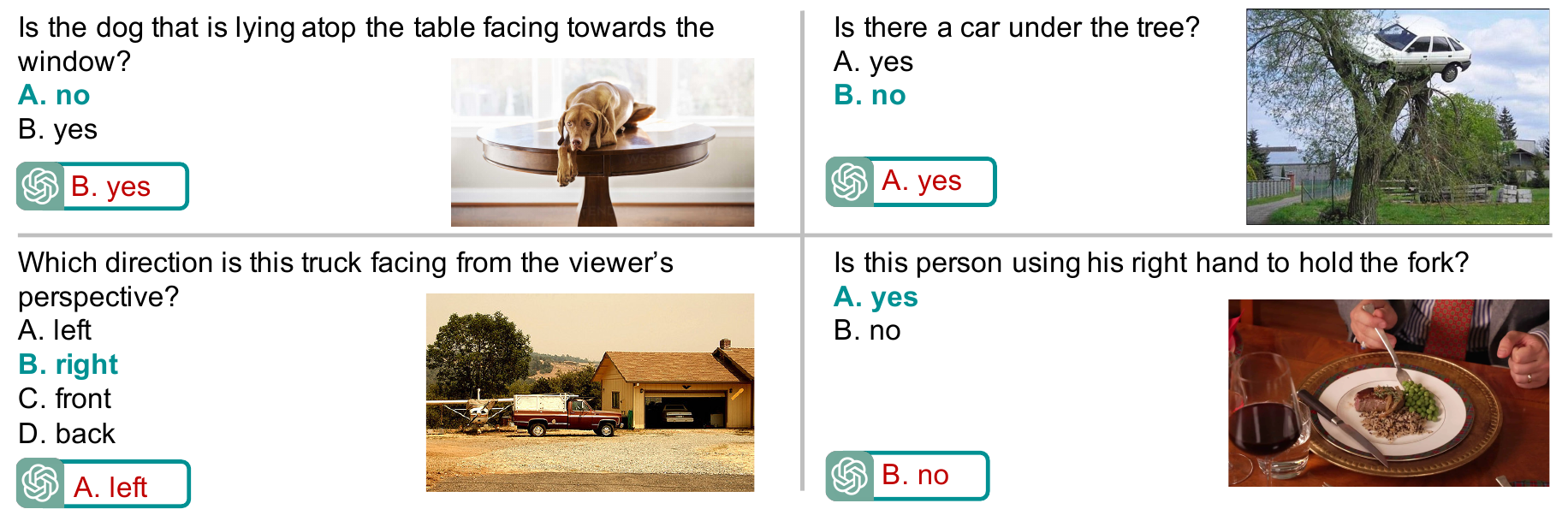}}
    \caption{Instances are identified where the spatial reasoning capabilities of GPT-4V \cite{gpt4} fall short (Date accessed: June 6, 2024) due to inaccurate spatial understanding. Text in \textcolor{red}{red} signifies an incorrect response. All the images referenced are from our Spatial-MM benchmark which are sourced from Internet.}
    \label{fig:examples}
\end{figure*}

%% file: figures/piplibe.tex
\begin{figure*}[h!]
    \centering{
    \includegraphics[width=1.0\textwidth]{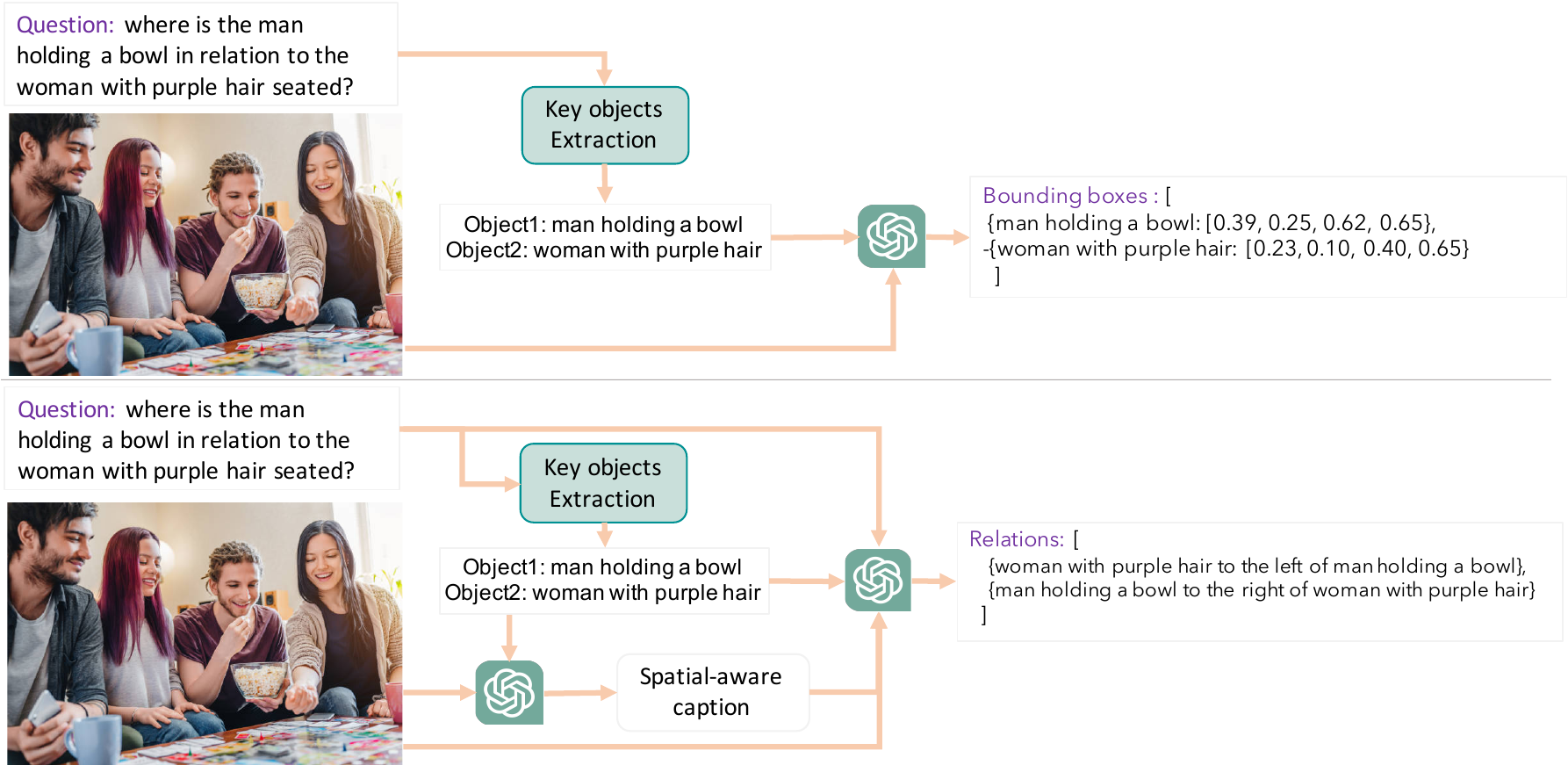}}
    \caption{The proposed pipelines for Bounding boxes generation (top) and Scene graph generation (bottom).}
    \label{fig:pieline}
\end{figure*}

%% file: figures/bbx_visulization.tex
\begin{figure}[h]
    \centering
    \resizebox{.5\textwidth}{!}{
    \includegraphics{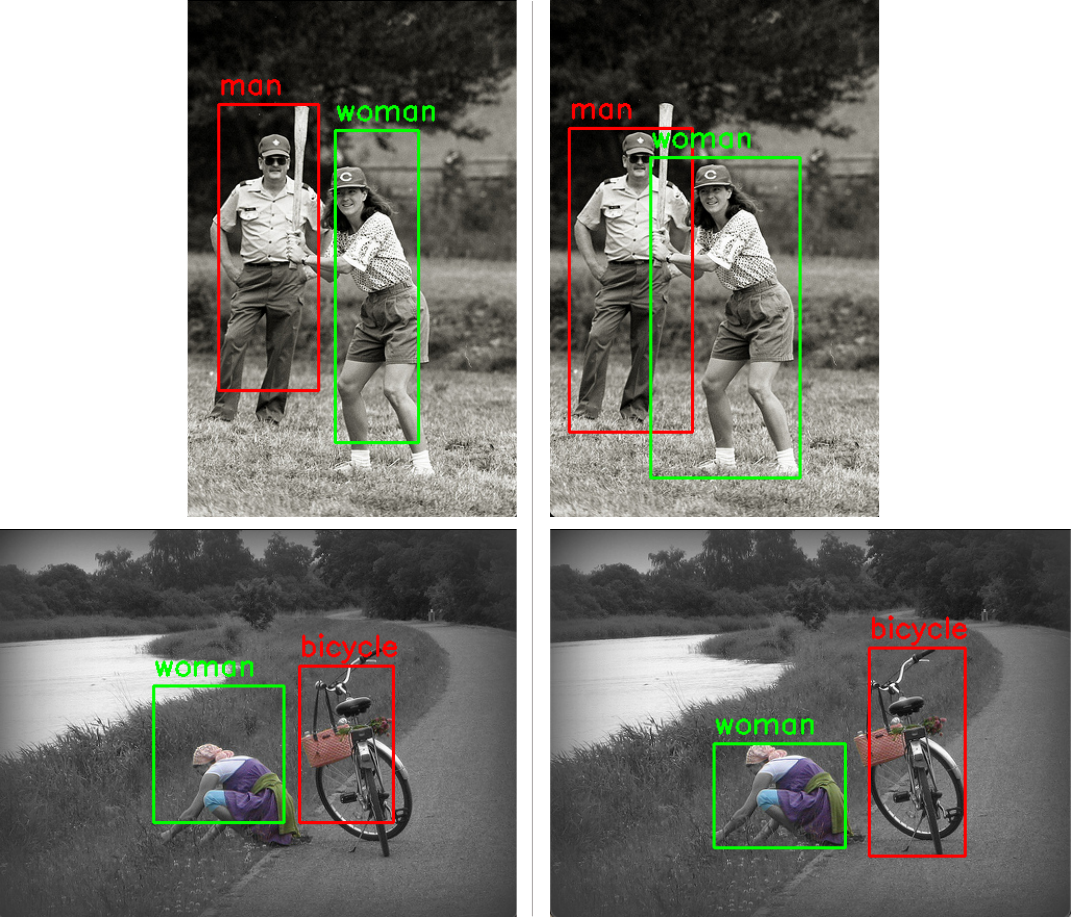}}
    \caption{A comparison between the ground truth bounding boxes from GQA dataset on the left, and the synthesized bounding boxes by GPT-4o on the right.}
    \label{fig:bbx}
\end{figure}

%% file: figures/spatial_rel_distribution.tex
\begin{figure*}[h!]
    \centering
    \resizebox{.99\textwidth}{!}{
    \includegraphics{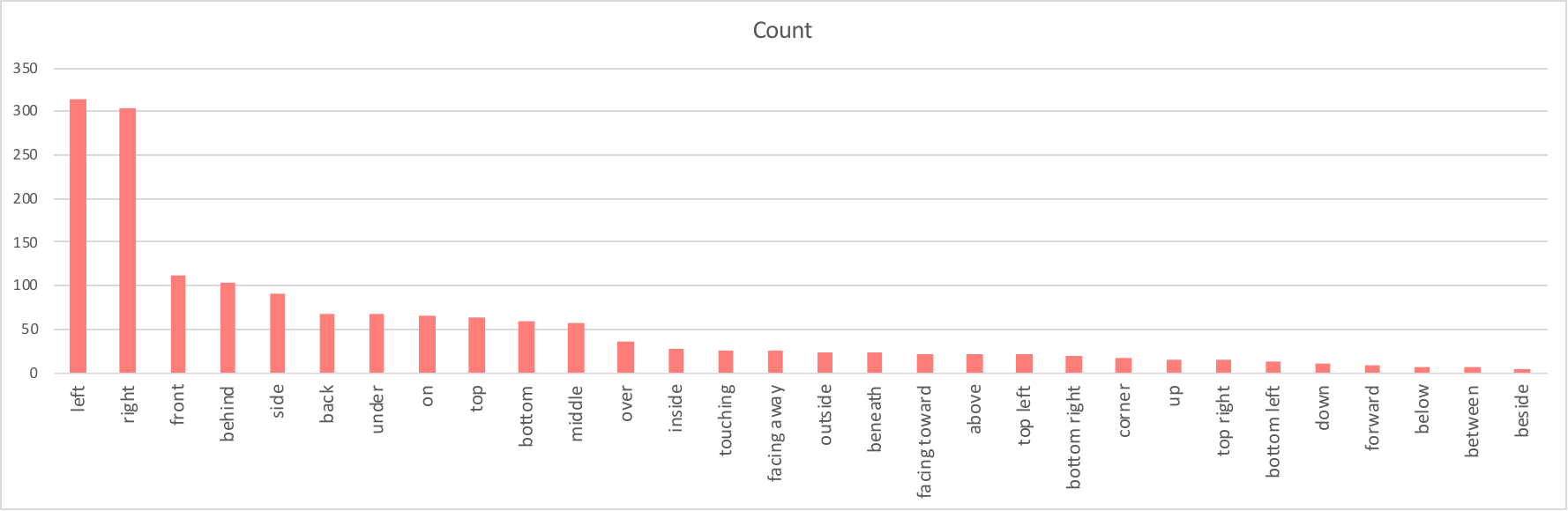}}
    \caption{Relation distribution of Spatial-MM dataset(sorted by frequency). Top 30 most frequent relations are included.}
    \label{fig:rel_distribution}
\end{figure*}

%% file: tables/preposition_analyses.tex
\begin{table}
\resizebox{0.99\linewidth}{!}{
\begin{tabular}{l|cccc}
\hline

Prepositions  & All & front-behind & left-right & top-bottom\\ \hline
Number of instances & 1451  & 26 & 1032  & 393 \\ \hline
Gemini (stan) & 18.52 & 15.38 & 19.16 & 17.05  \\
Gemini (stan + bbox) & 34.56 & 44.00 & 38.60 & 23.41  \\
Gemini (stan + SG)  & 45.24 & 60.00 & 49.36 & 33.59  \\
Gemini (stan + SG + bbox) & 41.00 & 36.00 & 44.64 & 31.81  \\ \hline
GPT-4v (stan)  & 23.71 & 79.17 & 18.18 & 63.10  \\
GPT-4v (stan + bbox) & 41.58 & 62.50 & 39.39 & 75.57  \\
GPT-4v (stan + SG) & \textbf{77.32} & \textbf{95.83} & \textbf{75.38} & \textbf{100.00} \\ \hline
LLaVA (stan) & 29.57 & 19.23 & 33.43 & 20.10  \\
LLaVA (stan + bbox) & 30.53 & 23.08 & 33.62 & 22.90  \\
LLaVA (stan + SG)  &41.83 &46.15 & 48.93&22.90 \\
LLaVA (stan + SG + bbox) & 38.73 &46.15 &44.38 & 23.41 \\ \hline
\end{tabular}}
\caption{Results categorized by different spatial prepositions on GQA-spatial dataset. We divided them by opposite preposition pairs: \textit{front-behind}, \textit{left-right}, and \textit{top-bottom}.}
\label{tab:preposition-analyses}
\end{table}